\pdfoutput=1
\documentclass[11pt]{article}

\usepackage[T1]{fontenc}
\usepackage[utf8]{inputenc}
\usepackage{lmodern}
\usepackage{microtype}
\usepackage[margin=1in]{geometry}
\usepackage{authblk}
\usepackage{graphicx}
\usepackage{amsmath,amssymb}
\usepackage{booktabs,array}
\usepackage{cite}
\usepackage{xurl}
\usepackage{caption}
\usepackage[hidelinks]{hyperref}

\newcommand{\TOneN}{T1n}
\newcommand{\TOneC}{T1c}
\newcommand{\TTwoW}{T2w}
\newcommand{\TTwoF}{T2-FLAIR}
\newcommand{\selected}{T1c+\TTwoF}

\title{Partial Information Decomposition as a Multi-Contrast 3D MRI Selection Strategy for Resource-Constrained Deep Neural Network Training in Brain Tumor Segmentation}
\author[1]{Agamdeep S. Chopra\thanks{ORCID: \href{https://orcid.org/0009-0003-4386-7894}{0009-0003-4386-7894}}}
\author[1]{Mehmet Kurt\thanks{ORCID: \href{https://orcid.org/0000-0002-5618-0296}{0000-0002-5618-0296}}}
\affil[1]{Department of Mechanical Engineering, University of Washington, Seattle, WA, USA}
\affil[ ]{\texttt{\{achopra4,mkurt\}@uw.edu}}
\date{}

\begin{document}
\maketitle

\begin{abstract}
Multi-contrast 3D MRI segmentation can be computationally demanding when all available sequences are used. We evaluate a pre-training Partial Information Decomposition framework that ranks input pairs according to their redundant, unique, and synergistic information about regional tumor burden and selects the highest-ranked pair for downstream training. Applied to \TOneN, \TOneC, \TTwoW, and \TTwoF{} MRI, the framework selected \selected. We then trained eleven architecturally identical lightweight 3D U-Nets using different input configurations. On an independent test cohort, \selected{} was the strongest two-input configuration and ranked second overall in mean Dice (0.676 versus 0.687 for all four inputs). Independent Shapley analysis on the full-input model also identified \TTwoF{} and \TOneC{} as the most influential inputs and their pairwise interaction as the strongest. These findings demonstrate the practical value of PID based pre-training selection for identifying compact, informative MRI input sets before costly 3D model development.
\end{abstract}

\noindent\textbf{Keywords:} Partial information decomposition, MRI sequence selection, Brain tumor segmentation, Resource-constrained learning
\vspace{0.75em}

\section{Introduction}
Multi-contrast MRI supports brain tumor segmentation because each sequence emphasizes complementary tissue characteristics. T1-weighted MRI depicts anatomy, contrast enhanced T1 highlights enhancing tumor, T2 MRI is sensitive to fluid, and T2-FLAIR suppresses cerebrospinal fluid to emphasize edema and infiltrative abnormalities. Consequently, brain tumor benchmarks and high-performing 3D segmentation frameworks commonly integrate all available contrasts for multi-region segmentation~\cite{menze2015brats,corebt2026,cicek2016unet3d,isensee2021nnunet}.

Although, this approach can be expensive in computationally constrained settings. Each additional sequence increases storage, preprocessing, data-transfer, memory, and training costs, which are amplified in 3D by large patch and volume tensors and by repeated experiments across folds, architectures, and hyperparameter settings. The practical question is therefore whether a smaller input subset can be selected before substantial resources are committed to downstream model development.

Standard subset selection relies on either exhaustive model training or single-source relevance measures. Exhaustive evaluation provides direct task performance estimates but scales poorly and must be repeated for each downstream architecture. Mutual information based methods offer a cheaper alternative by balancing target relevance against redundancy~\cite{peng2005mrmr}, but they may overlook synergy, whereby two sequences are jointly informative even when neither is individually dominant.

\begin{figure}[htbp]
\centering
\includegraphics[width=0.35\textwidth]{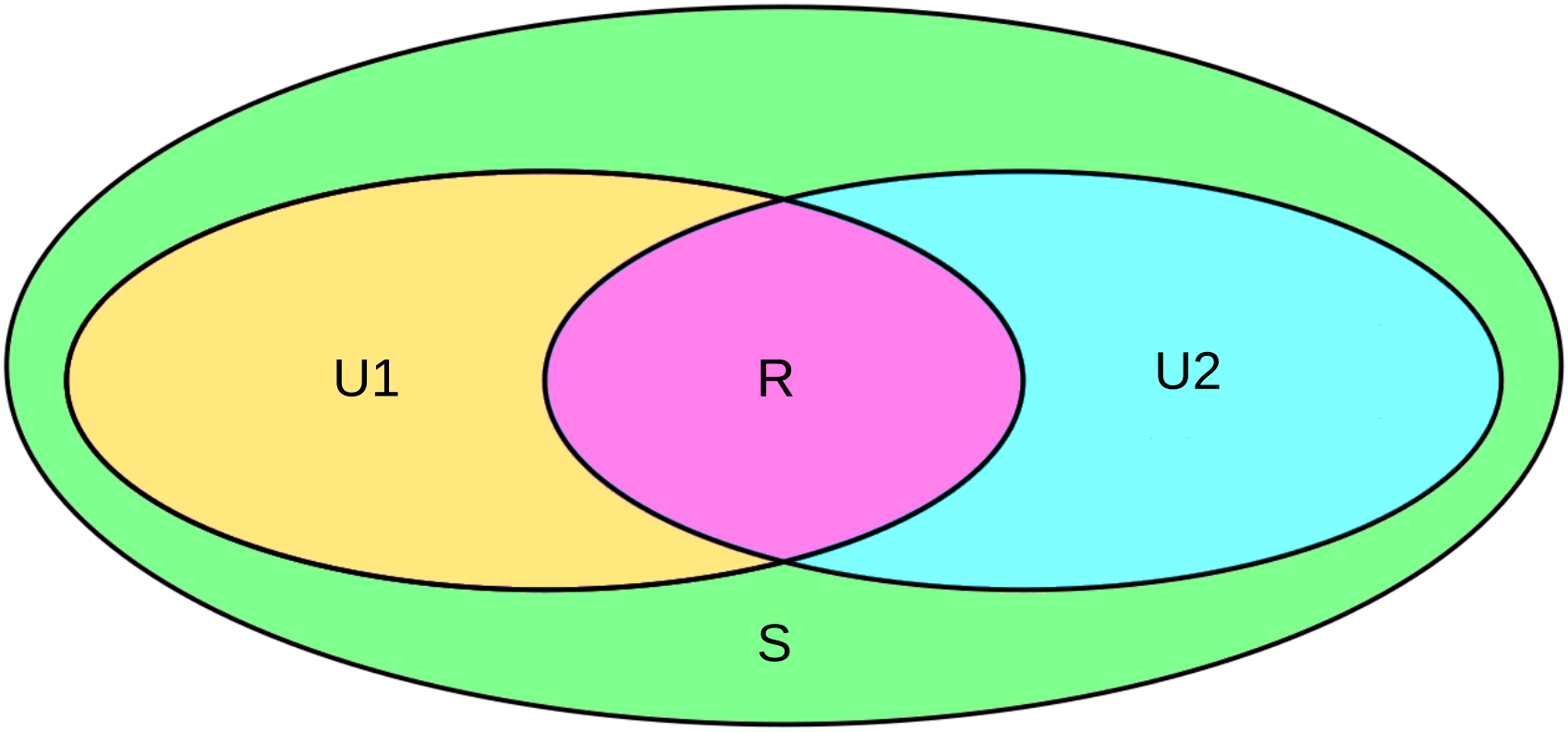}
\caption{Diagram representing partial information decomposition for two sources, $X_1$ and $X_2$, relative to a target $Y$. The total information is partitioned into source-specific contributions, $U_1$ and $U_2$, redundant information $R$ provided by both sources, and synergistic information $S$ available only from their joint observation.}
\label{fig:pid_schematic}
\end{figure}

Partial information decomposition (PID) separates target information into redundant, unique, and synergistic components~\cite{williams2010pid,bertschinger2014unique}. It has been used to define feature relevance and guide feature selection~\cite{wollstadt2021feature,westphal2024pidf}
, while recent works such as SFL-Net applied PID-guided latent factorization to multi-contrast MRI to PET synthesis~\cite{chopra2026sflnetsourcefactorizedlatentrepresentation}
. Building on this work, prior studies of MRI sequence contributions in brain tumor segmentation~\cite{zhou2021feature,sadegheih2024modality}, and contrast-level Shapley attribution~\cite{ren2025shapley}, we introduce a general $n$-to-2 pre-training framework for multi-contrast MRI tumor segmentation. A compact autoencoder is trained for each input sequence, the resulting embeddings are quantized, and all $\binom{n}{2}$ pairs are ranked using a region-aware minimum mutual information PID (MMI-PID) definition score~\cite{barrett2015mmi}. The highest scoring pair is then selected for downstream model development. We evaluate the framework using four CoRe-BT MRI sequences~\cite{corebt2026}, comparing the full four-input model, all six two-input combinations, and four single-input models with patient-level statistical testing and Shapley attribution under a restricted compute budget.

\section{Method}
\subsection{PID Pair Selection}

\begin{figure}[htbp]
\centering
\includegraphics[width=0.95\textwidth]{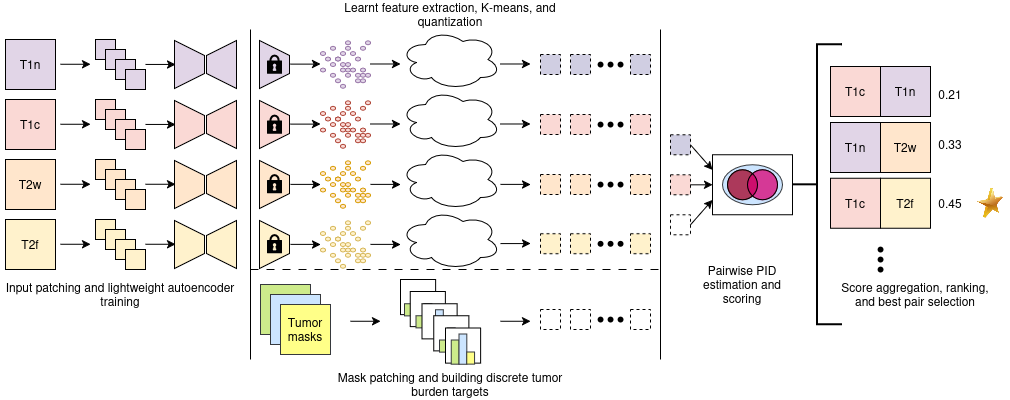}
\caption{PID pair-selection pipeline. Spatially aligned patches
from each MRI sequence are encoded using sequence-specific autoencoders.
The frozen latent representations are quantized into discrete source codes,
while aligned tumor masks provide discretized regional tumor-burden targets.
All two-input combinations are evaluated using region-aware PID, ranked by
their aggregate scores, and the highest-ranked pair is retained for
downstream model training.}
\label{fig:pipeline}
\end{figure}

Figure~\ref{fig:pipeline} summarizes the selection process. For each training
subject, we sampled 128 spatially corresponding
$32\times32\times32$ patches from all MRI sequences and the tumor mask. 70\%
of patch centers were drawn from the tumor foreground. For each tumor region
$c \in \{\text{edema},\text{enhancing tumor},\text{necrotic}\}$,
tumor burden was defined as the fraction of patch voxels assigned to that
region. Zero-burden patches formed class 0, while positive burdens were divided
into three training-set quantile classes.

A separate shallow 3D autoencoder was trained for each MRI sequence to
reconstruct its input patches using mean-squared error. The encoder comprised
three $3\times3\times3$ convolutional blocks with stride two and 8, 16, and
32 output channels respectively. Each convolution was followed by
instance normalization and a ReLU activation. For a
$32\times32\times32$ input patch, the encoder produced a
$32\times4\times4\times4$ feature map, which was flattened and linearly
projected to a 32-dimensional latent vector.
After training, the encoders were frozen and the decoders were discarded.
For each sequence $m$, a separate $K$-means model with $K=16$ was fitted to
the training latent vectors, and each cluster assignment defined the discrete
source code $q_m$~\cite{hinton2006autoencoder,lloyd1982kmeans}.

For every input pair $(a,b)$ and tumor region $c$, mutual information was
estimated from the empirical joint distribution of the two source codes and
the discretized tumor-burden target $T_c$~\cite{cover2006information}. Following the bivariate PID framework~\cite{williams2010pid}, we decomposed
the information that each pair of quantized source codes provided about
the regional tumor-burden target. We used the MMI formulation as a computationally simple proxy for bivariate PID~\cite{barrett2015mmi}. For input pair $(a,b)$ and
tumor region $c$, the PID terms were
\begin{align}
R_{ab}^{c}
&=
\min\!\left\{I(q_a;T_c),I(q_b;T_c)\right\},
\nonumber\\
U_{a|b}^{c}
&=
I(q_a;T_c)-R_{ab}^{c},
\qquad
U_{b|a}^{c}
=
I(q_b;T_c)-R_{ab}^{c},
\nonumber\\
S_{ab}^{c}
&=
I((q_a,q_b);T_c)
-
R_{ab}^{c}
-
U_{a|b}^{c}
-
U_{b|a}^{c}.
\end{align}

To rank the candidate pairs, we defined an aggregate score inspired by PID-based feature selection criteria that favor informative and non-redundant feature sets~\cite{wollstadt2021feature,westphal2024pidf}:
\begin{equation}
\mathrm{Score}(a,b)
=
\sum_c w_c
\left[
I((q_a,q_b);T_c)
+
\lambda
\left(
U_{a|b}^{c}
+
U_{b|a}^{c}
+
S_{ab}^{c}
\right)
\right],
\qquad \lambda=0.5,
\end{equation}
where $w_c$ was the normalized inverse frequency of positive patches for region $c$. Thus, joint target information determined the baseline pair score, while unique and synergistic information received an additional weight relative to redundant information. The selected pair was
\begin{equation}
\mathcal{M}^{*}
=
\arg\max_{\{a,b\}\subset\mathcal{M}}
\mathrm{Score}(a,b),
\qquad
|\mathcal{M}^{*}|=2.
\end{equation}

\subsection{Lightweight 3D U-Net}
The downstream segmentation model was a lightweight 3D U-Net with three resolution levels~\cite{cicek2016unet3d}. Each encoder and decoder block contained two $3\times3\times3$ convolutions followed by ReLU activations and Instance Normalization. Max pooling was used for downsampling, while nearest-neighbor interpolation followed by $1\times1\times1$ convolution was used for upsampling. A final $1\times1\times1$ convolution produced the output logits.

Models were trained using $64\times64\times64$ patches, a batch size of one, and 256 patches per epoch. The training objective combined Dice loss and binary cross-entropy, balancing region overlap with voxelwise supervision. Optimization used AdamW with a weight decay of $10^{-4}$~\cite{loshchilov2019adamw}. The learning rate was reduced from $10^{-3}$ to $10^{-5}$ using a cosine schedule over 200 epochs~\cite{loshchilov2017sgdr}. A 30 subject validation cohort was used for development stage benchmarking, and final evaluation was performed on an independent held-out 28 subject test cohort.

\section{Experiments}
We used 132 training, 30 validation, and 28 test subjects from the MRI and tumor-mask component of CoRe-BT dataset~\cite{corebt2026}. Subjects missing any required sequence were excluded. Inputs were \TOneN, \TOneC, \TTwoW, and \TTwoF. Reported targets were edema, enhancing tumor, and necrotic/non-enhancing tumor. Volumes were cropped, clipped to the 0.5th--99.5th percentile range, and normalized for training.

PID selected \selected{} as the highest-ranked input pair for this task. We then trained 11 architecturally identical 3D U-Net models (one four-input model, all six pairwise models, and four single-input models). This design tested whether the PID ranking predicted downstream segmentation performance.

The models were evaluated on the held-out set using patient-level macro-averaged Dice,
HD95, sensitivity, and precision across tumor regions. Statistical analyses
included 10,000 bootstrap resamples, Spearman correlation, paired Wilcoxon
tests, Friedman tests with Holm correction, and a $-0.03$ Dice
non-inferiority margin~\cite{holm1979multiple}. Shapley values over all
16 input coalitions were computed for the four-input model, replacing omitted
inputs with training set channel means inside the brain mask and zeros outside
~\cite{shapley1953value,ren2025shapley}.

The complete experiment used one RTX 3090 with GPU memory capped at 2~GB, four logical CPU cores, and 4~GB of system memory for training and feature extraction.

\section{Analysis and Results}
\subsection{PID Selection and Segmentation}
PID ranked \selected{} first with a score of 0.963. Across the six pair models, the PID score was positively associated with test-set Dice ($\rho=0.600$, $p=0.208$) and negative HD95 ($\rho=0.600$, $p=0.208$). Both associations were directionally consistent with downstream performance, but neither was statistically significant, and the analysis had limited power because only six pairs were available.

\begin{table}[htbp]
\centering
\caption{Region-specific PID scores for all input-pair combinations. Nec., Ed., and Enh. denote necrotic tumor, edema, and enhancing tumor, respectively.}
\label{tab:pid_region_scores}
\small
\setlength{\tabcolsep}{5pt}
\begin{tabular}{lrrrrr}
\toprule
Input pair & Nec. & Ed. & Enh. & Aggregate & Rank \\
\midrule
\textbf{\TOneC+\TTwoF}
& \textbf{0.789}
& \textbf{1.106}
& \textbf{1.032}
& \textbf{0.963}
& \textbf{1} \\

\TOneC+\TTwoW
& 0.785
& 0.777
& 0.947
& 0.841
& 2 \\

\TOneN+\TTwoF
& 0.745
& 0.932
& 0.736
& 0.793
& 3 \\

\TTwoW+\TTwoF
& 0.669
& 0.849
& 0.651
& 0.712
& 4 \\

\TOneN+\TTwoW
& 0.662
& 0.680
& 0.590
& 0.642
& 5 \\

\TOneN+\TOneC
& 0.455
& 0.409
& 0.728
& 0.540
& 6 \\
\bottomrule
\end{tabular}
\end{table}

On the test cohort, the full model achieved average Dice 0.687 and HD95 16.09. \selected{} achieved Dice 0.676 and HD95 16.77, making it the strongest pair and second overall (Table~\ref{tab:results}). The selected pair retained 98.5\% of the full-model mean Dice while using half the input channels.

\begin{table}[htbp]
\centering
\caption{Test-set performance for the top input configurations. Ed, En,
and Ne denote edema, enhancing tumor, and necrotic tumor, respectively.
HD95, average Dice, sensitivity, and precision are patient-level macro
averages across the three regions.}
\label{tab:results}
\small
\setlength{\tabcolsep}{2.5pt}
\renewcommand{\arraystretch}{1.05}
\begin{tabular}{lrrrrrrrr}
\toprule
Inputs & Ch. & HD95 & Avg. Dice & Sens. & Prec. &
Dice Ed & Dice En & Dice Ne \\
\midrule
All four
& 4 & \textbf{16.09} & \textbf{0.687}
& \textbf{0.709} & \textbf{0.714}
& 0.550 & \textbf{0.800} & \textbf{0.710} \\

\textbf{\selected}
& 2 & 16.77 & 0.676
& \textbf{0.709} & 0.699
& \textbf{0.570} & 0.767 & 0.692 \\

\TOneC+\TTwoW
& 2 & 18.06 & 0.650
& 0.670 & 0.683
& 0.520 & 0.721 & 0.707 \\

\TOneC
& 1 & 21.71 & 0.589
& 0.606 & 0.622
& 0.534 & 0.574 & 0.658 \\

\TOneN+\TOneC
& 2 & 18.21 & 0.587
& 0.594 & 0.612
& 0.488 & 0.613 & 0.661 \\
\bottomrule
\end{tabular}
\end{table}

The selected vs. full Dice difference was $-0.010$ (95\% CI $[-0.052,0.034]$; paired Wilcoxon $p=0.0505$). HD95 increased by 0.68 (95\% CI $[-4.54,5.09]$; $p=0.174$). Sensitivity was nearly identical (difference $<0.001$; $p=0.991$), and precision differed by $-0.015$ ($p=0.099$). Dice non-inferiority was not established given our choice of margin.

Model configuration affected Dice ($\chi^2_{10}=164.63$, $p<10^{-29}$, Kendall's $W=0.588$) and HD95 ($\chi^2_{10}=79.07$, $p<10^{-12}$, $W=0.282$). The selected pair significantly outperformed most lower-performing pairwise and single-input models after Holm correction. However, differences between the selected pair and either the full model or \TOneC+\TTwoW{} were not significant after correction.

\subsection{Post-hoc Attribution}
\begin{figure}[htbp]
\centering
\includegraphics[width=0.98\textwidth]{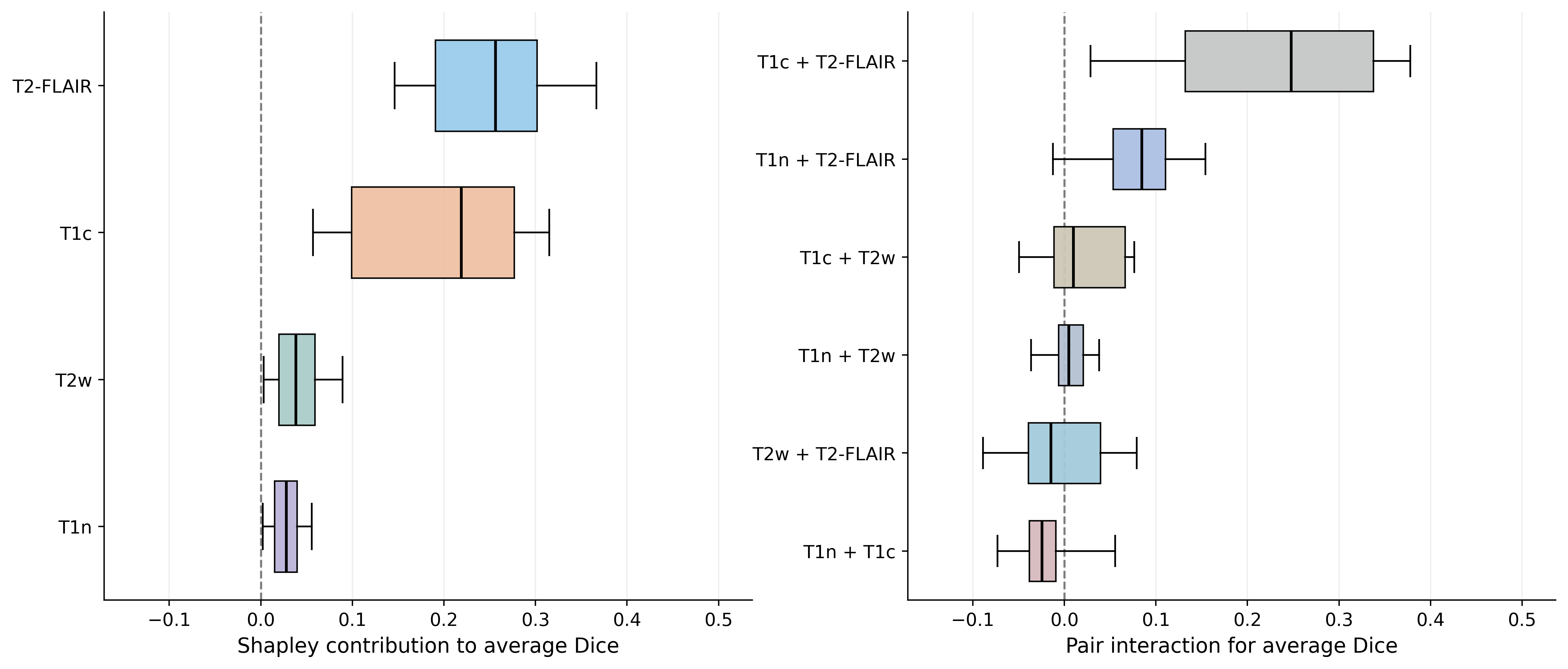}
\caption{Test-set Shapley analysis of the four-input model. \TTwoF{} and \TOneC{} had the largest input contributions, and their pairwise interaction was the strongest.}
\label{fig:shapley}
\end{figure}
Shapley analysis ranked \TTwoF{} first (0.336; 95\% CI 0.298--0.374) and \TOneC{} second (0.258; 95\% CI 0.203--0.311). Their interaction was also largest (0.307; 95\% CI 0.241--0.373; Fig.~\ref{fig:shapley}). Pair interactions differed across the six combinations ($\chi^2_5=71.22$, $p<10^{-13}$, Kendall's $W=0.509$). This independent analysis agreed with the PID-selected input pair.

\section{Discussion}
\subsection{PID Identified the Strongest Input Pair}
The central result is that the MMI-PID score selected the strongest two-input configuration before segmentation training. \selected{} ranked first by the information score and was independently identified as the best pair in the 11-model evaluation. This agreement supports the use of the proposed score as a pre-training input pair selector.

The selected model used half the input channels and retained 98.5\% of the full input model mean Dice. Paired tests did not provide clear evidence of differences in Dice, HD95, sensitivity, a larger study is required to reliably draw conclusions. However, our work only serves as a proof of concept and is limited by only 4 input contrasts and small dataset splits.

\subsection{The Selected Pair Matches the Imaging Task}
The selection of \TOneC{} and \TTwoF{} is consistent with the reported regions. \TOneC{} directly emphasizes enhancing tumor, while \TTwoF{} is sensitive to edema and surrounding tissue abnormality. Their combination therefore covers two distinct components of the segmentation target. The region-specific PID scores reflected this pattern as  (Table~\ref{tab:pid_region_scores}). \selected{} achieved the highest score for all three evaluated regions, including edema, enhancing tumor, and necrotic tumor.

Shapley attribution provided a separate model-based audit. \TTwoF{} and \TOneC{} had the two largest individual contributions, and their interaction was larger than every other pair interaction. However, PID and Shapley address different questions. PID ranks inputs before downstream training, whereas Shapley measures how a specific trained full model uses its inputs under a defined missing-input baseline. Their agreement strengthens the interpretation that \selected{} captures both strong individual signal and useful cross-input dependence.

\subsection{Implications for Resource-Constrained Training}
Selection occurs before costly segmentation training. For $n$ candidate inputs,
the method trains $n$ shallow autoencoders and evaluates $\binom{n}{2}$ pair
scores, avoiding a separate downstream model for every pair.

These pre-training steps are comparatively lightweight and reusable across
architectures. Reducing the number of inputs also lowers data-transfer,
preprocessing, latency, augmentation, memory, and computation costs.

\section{Conclusion}
In this work, our framework using shallow autoencoder encoding, latent quantization, and a region-aware MMI-PID score identified the strongest two-input configuration before segmentation training. The framework selected \selected, which halved the number of input channels, retained 98.5\% of the full-model test Dice, and was independently supported by Shapley attribution. This study provides a focused proof of concept for PID based input selection as a practical and interpretable pre-training strategy for reducing multi-contrast 3D input complexity. Future work should evaluate its robustness in larger cohorts and settings with more than four candidate inputs.

\bibliographystyle{unsrt}
\bibliography{references}

\end{document}